\documentclass[sigconf]{acmart}

\usepackage{multirow}
\usepackage{multicol}
\usepackage{colortbl}
\usepackage{diagbox}

\usepackage{algorithm}
\usepackage{algpseudocode}

\AtBeginDocument{%
	\providecommand\BibTeX{{%
			\normalfont B\kern-0.5em{\scshape i\kern-0.25em b}\kern-0.8em\TeX}}}

\begin{document}

	\newcommand{\Nk}{{N}^{k}}
	\newcommand{\Mk}{{M}^{k}}
	\newcommand{\Lk}{{L}^{k}}
	\newcommand{\yk}{{y}^{k}}
	\newcommand{\bfxk}{{\bf{x}}^{k}}
	\newcommand{\bfPk}{{\bf{P}}^{k}}
	\newcommand{\bfMk}{{\bf{M}}^{k}}
	\newcommand{\bfWk}{{\bf{W}}^{k}}
	\newcommand{\bfWEk}{{\bf{W}}_{E}^{k}}
	\newcommand{\bfhatWEk}{{\bf{\hat{W}}}_{E}^{k}}
	\newcommand{\bftildeWEk}{{\bf{\tilde{W}}}_{E}^{k}}
	\newcommand{\bfWTk}{{\bf{W}}_{T}^{k}}
	\newcommand{\bfWEPk}{{\bf{W}}_{EP}^{k}}
	\newcommand{\bfWESk}{{\bf{W}}_{ES}^{k}}
	
	\newcommand{\bfuk}{{\bf{u}}^{k}}
	\newcommand{\bfUk}{{\bf{U}}^{k}}
	\newcommand{\bfvk}{{\bf{v}}^{k}}
	\newcommand{\ovbfvk}{{\overline{\bf{v}}}^{k}}

	\newcommand{\tPk}{{\theta_P^k}}
	\newcommand{\tEk}{{\theta_E^k}}
	\newcommand{\hattEk}{{\hat{\theta}_E^k}}
	\newcommand{\ovtEk}{{\overline{\theta}_E^k}}
	\newcommand{\tGk}{{\theta_G^k}}
	\newcommand{\tHk}{{\theta_H^k}}
	
	\newcommand{\tFk}{{\theta_F^k}}
	\newcommand{\tTk}{{\theta_T^k}}
	
	\newcommand{\calLk}{{\mathcal{L}}^{k}}
	\newcommand{\calDk}{{\mathcal{D}}^{k}}
	\newcommand{\calVk}{{\mathcal{V}}^{k}}
	
	\newcommand{\bfek}{{\bf{e}}^{k}}
	\newcommand{\bfhk}{{\bf{h}}^{k}}
	\newcommand{\bfok}{{\bf{o}}^{k}}
	\newcommand{\bfgk}{{\bf{g}}^{k}}
	
	\newcommand{\bfx}{{\bf{x}}}
	\newcommand{\bfg}{{\bf{g}}}
	\newcommand{\bfP}{{\bf{P}}}
	\newcommand{\bfW}{{\bf{W}}}
	\newcommand{\bfM}{{\bf{M}}}
	\newcommand{\bfWE}{{\bf{W}}_{E}}
	\newcommand{\bfWT}{{\bf{W}}_{T}}
	\newcommand{\bfWEP}{{\bf{W}}_{EP}}
	\newcommand{\bfWES}{{\bf{W}}_{ES}}
	
	\newcommand{\bfK}{{\bf{K}}}
	\newcommand{\bfH}{{\bf{H}}}
	\newcommand{\bfL}{{\bf{L}}}
	
	\newcommand{\tE}{{\theta_E}}
	\newcommand{\tG}{{\theta_G}}
	\newcommand{\tH}{{\theta_H}}
	\newcommand{\tP}{{\theta_P}}
	
	\newcommand{\tF}{{\theta_F}}
	\newcommand{\tT}{{\theta_T}}
	
	\newcommand{\calL}{{\mathcal{L}}}
	\newcommand{\calD}{{\mathcal{D}}}
	\newcommand{\calV}{{\mathcal{V}}}
	\newcommand{\calP}{{\mathcal{P}}}
	
	\newcommand{\calR}{{\mathcal{R}}}
	\newcommand{\calI}{{\mathcal{I}}}
	\newcommand{\calO}{{\mathcal{O}}}
	
	\newtheorem{remark}{Remark}
	\newtheorem{mydef}{Definition}
	
	\title{Preliminary Steps Towards Federated Sentiment Classification}

	
	\author{Xin-Chun Li, Lan Li, De-Chuan Zhan}
	\affiliation{%
		\institution{State Key Laboratory for Novel Software Technology, Nanjing University}
	}
	\email{{lixc, lil}@lamda.nju.edu.cn, zhandc@nju.edu.cn}
	
	\author{Yunfeng Shao, Bingshuai Li, Shaoming Song}
	\affiliation{%
		\institution{Huawei Noah's Ark Lab}
	}
	\email{{shaoyunfeng, libingshuai, shaoming.song}@huawei.com}
	
	
	
	\begin{abstract}
		Automatically mining sentiment tendency contained in natural language is a fundamental research to some artificial intelligent applications, where solutions alternate with challenges.
		Transfer learning and multi-task learning techniques have been leveraged to mitigate the supervision sparsity and collaborate multiple heterogeneous domains correspondingly.
		Recent years, the sensitive nature of users' private data raises another challenge for sentiment classification, i.e., data privacy protection. 
		In this paper, we resort to federated learning for multiple domain sentiment classification under the constraint that the corpora must be stored on decentralized devices. In view of the heterogeneous semantics across multiple parties and the peculiarities of word embedding, we pertinently provide corresponding solutions.
		First, we propose a Knowledge Transfer Enhanced Private-Shared (KTEPS) framework for better model aggregation and personalization in federated sentiment classification.
		Second, we propose KTEPS$^\star$ with the consideration of the rich semantic and huge embedding size properties of word vectors, utilizing Projection-based Dimension Reduction (PDR) methods for privacy protection and efficient transmission simultaneously.
		We propose two federated sentiment classification scenes based on public benchmarks, and verify the superiorities of our proposed methods with abundant experimental investigations.
	\end{abstract}

	
	
	
	\maketitle
	
	\section{Introduction}
	Sentiment Classification (SC)~\cite{SC-DeepSurvey-2018} is a fundamental task for capturing and understanding users' emotions in natural language, which has raised pervasive attention and found wide applications. The goal of SC is to identify the overal sentiment polarity of a document, which is a special case of text classification. Researchers have studied various SC settings, such as document-level SC~\cite{DocSC-CNN-NAACL-HTL2015,DocSC-RNN-EMNLP2015}, sentence-level SC~\cite{SentSC-MIL-TACL2018,SentSC-MLP-Financial-EMNLP2017} and aspect-level SC~\cite{AspectSC-AdaRNN-ACL2014,AspectSC-Hier-EMNLP2016},  which mainly differ in the granularity of the sentiment. Simultaneously, recent advances in deep neural network (DNN) have facilitated the success of SC, e.g., CNNs~\cite{DocSC-CNN-NAACL-HTL2015}, RNNs~\cite{DocSC-RNN-EMNLP2015}, memory networks~\cite{DocSC-UPMemory-EMNLP2017}, attention mechanisms~\cite{DocSC-UPAttn-EMNLP2016,DocSC-HAN-NAACL-HTL-2016}, transformer~\cite{SC-PivotTransformer-AAAI20}, etc.
	
	In the history of SC, challenges and solutions are always in alternation. Although DNNs have achieved huge success in SC, they are highly dependent on large annotated corpus. Additionally, the domain shift problem prevents a pre-trained model being directly applied to the target domain. To alleviate the dependence on large amounts of labeled data and mitigate the domain discrepancy, transfer learning (TL) methods are leveraged to provide effective solutions~\cite{SCL-EMNLP2006,Attn-CrossLingual-EMNLP2016,MVTL-ACL2020}. 
	A further challenge emerges when collaborating multiple corpus, where the semantic contexts vary a lot across domains. To guarantee acceptable performances for each domain, multi-task learning (MTL) techniques are fused into SC~\cite{CNN-MTL-ICML2008,MT-DNN-ACL2015,RNN-PS-IJCAI2016,ARC-Memory-EMNLP2016,DSR-NAACL2018,SADA-MTL-IJCAI2018,DAEA-IJCAI2019,ASP-MTL-ACL2017,MAN-NAACL-HLT2018}. Different multiple domain SC (MDSC) methods are proposed to extract domain-invariant knowledge across domains. 
	
	This is not an end. Higher requirements for data security~\cite{Secure-aggregation,DiffPrivate,DP-FL-WWW2020,Gboard} poses a new challenge for SC, i.e., data privacy protection. Traditional MDSC methods assume that text data from multiple domains can be shared without any risk of privacy leakage. However, the corpora used to train SC models often contain sensitive information of enterprises or customers in real-world applications, and leaking the private data may lead to responsibilities and risks. {\em What techniques can we introduce to solve the data privacy challenge? Are there any difficulties that we must overcome?}
	
	\begin{figure*}
		\includegraphics[width=\textwidth]{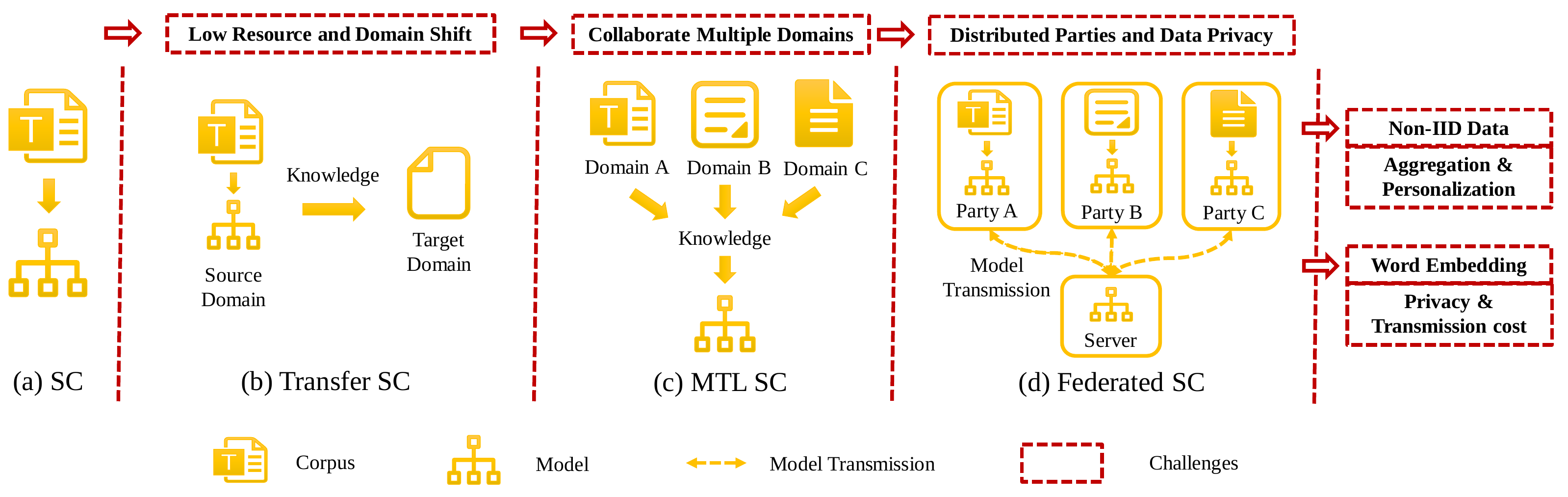}
		\caption{\small Illustration of challenges and solutions for Sentiment Classification (SC). (a) The training procedure of SC. (b) Transfer Learning (TL) methods are utilized to mitigate domain discrepancy and facilitate the low-resource target domain. (c) Multi-Task Learning (MTL) methods are incorporated to collaborate multiple domains. (d) Under the constraints of distributed parties and data privacy, we resort to Federated Learning (FL), where we must overcome two difficulties, i.e., the Non-IID data and Word Embedding problems.}
		\label{fig:teaser}
	\end{figure*}
	
	In this paper, we resort to federated learning (FL)~\cite{FedAvg,Fed-Concept,Fed-Advances,FedRS,FedPAN,FedPHP} for privacy-preserved MDSC. FL has been proposed as a natural solution for privacy protection. {\em Although FL has been applied to some language model applications (e.g., n-gram language model)~\cite{Gboard,FedAvg,Federated-ngram,Federated-OOV}, it has not yet been studied in SC as far as we know.} In fact, the existed methods simply combines FedAvg~\cite{FedAvg}, a classical FL framework, with NLP tasks without pertinent solutions. That is, they aggregate local models to a single global model without consideration of heterogeneous data distributions, i.e., the Non-IID problem~\cite{Fed-NonIID-Data,Fed-NonIID-Quagmire}. Additionally, word vectors are semantic rich and privacy sensitive representations (e.g., the gender bias)~\cite{WV-Gender-Bias-NeurIPS2016,WV-Debias-ACL2020}, which needs stricter privacy protection mechanisms. Furthermore, the huge embedding size also brings a transmission burden to FL. The whole motivation and the challenges to be solved in this paper are concluded in Figure.~\ref{fig:teaser}.
	
	We propose corresponding solutions for both Non-IID and word embedding challenges encountered in federated MDSC (FedMDSC). For Non-IID problems, we first present the correlations between MTL and FL methods when applied to MDSC from the aspect of DNN architecture design, i.e., the private-shared models~\cite{DSN-NeurIPS2016,ASP-MTL-ACL2017,DSR-NAACL2018,DAEA-IJCAI2019,FedPer-CoRR2019,FURL-CoRR2019,PFL-DA-CoRR2019,LG-FedAvg-CoRR2020}. Then, we propose a Knowledge Transfer Enhanced Private-Shared (KTEPS) framework for consideration of both global model aggregation and local model personalization simultaneously. In detail, local models are designed as DNNs with two parallel branches, where a task-specific classifier is kept private on local devices. A diversity term is added to implicitly separate domain invariant and domain specific information into two branches. To enhance the personalization ability of the private branch, we take advantage of knowledge distillation~\cite{KD-CoRR2015,DML-CVPR2018} to facilitate information flow from the global shared branch to the private one. For the word embedding problems, we utilize Projection-based Dimension Reduction (PDR) methods to reduce the size of word embeddings, which can simultaneously provide a stricter privacy protection mechanism and release the burden of communication in FL.
	
	To conclude, our contributions are listed as follows:
	\begin{itemize}
		\item We are the first to investigate the privacy-preserved MDSC as far as we know.
		\item We are not simply fusing FL into MDSC, and on the contrary, we provide specific solutions to overcome the fundamental difficulties, i.e., the Non-IID and word embedding problems.
		\item We sort out the relationship between MTL and FL methods from aspect of DNN architecture design, and propose KTEPS for better model aggregation and personalization.
		\item We utilize PDR methods to compress word embeddings for stricter privacy protection and lighter transmission.
		\item We construct two FedMDSC scenes based on public SC benchmarks and verify our proposed methods with abundant experimental studies.
	\end{itemize}
	
	\begin{figure*}
		\includegraphics[width=\textwidth]{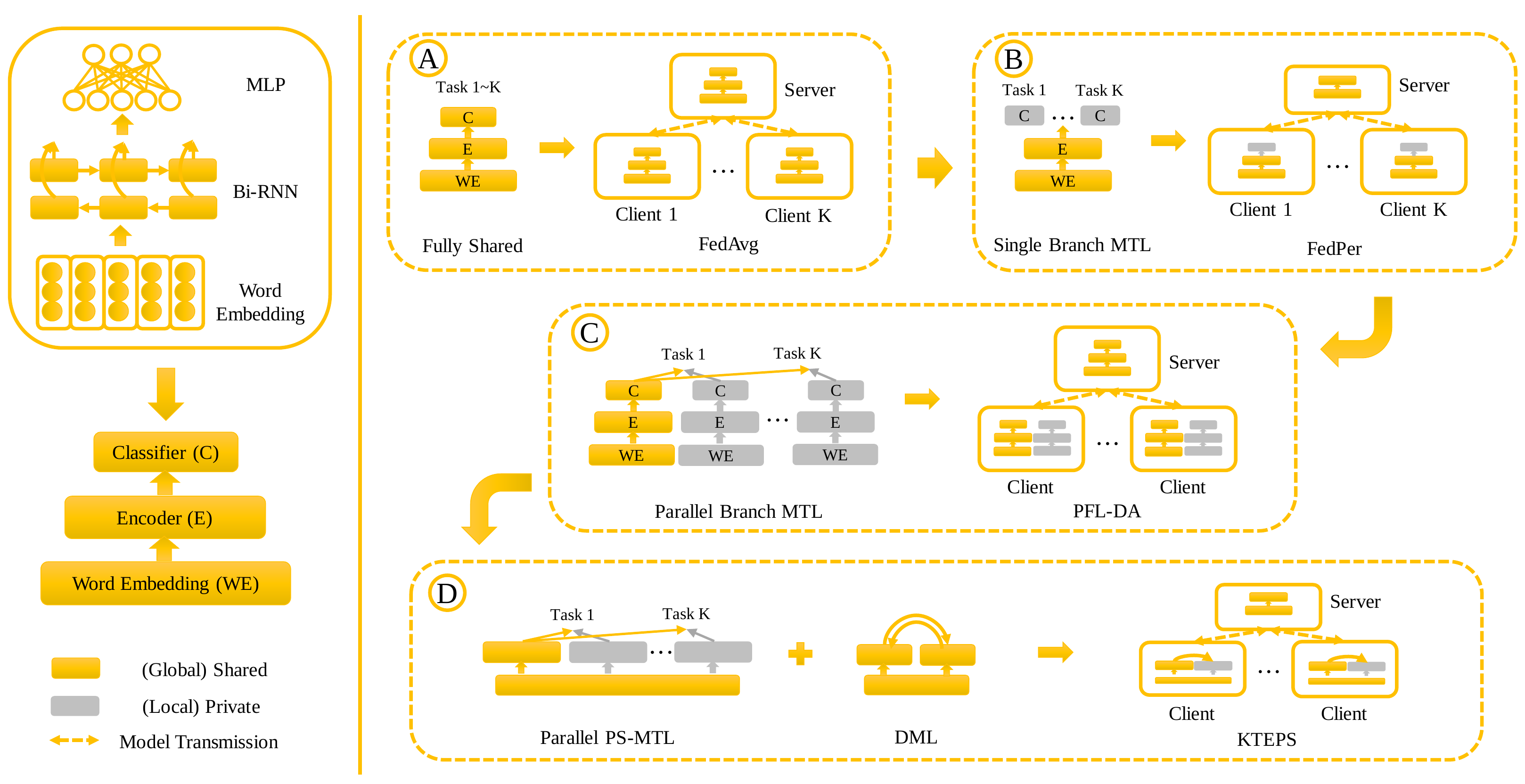}
		\caption{\small Illustration of correlations between MTL and FL methods when applied to MDSC. Left: the basic network architecture we utilize for SC. Right: (A) The fully-shared MTL and FedAvg for MDSC. (B) The single branch MTL and FedPer for MDSC. (C) The parallel branch MTL and PFL-DA for MDSC. (D) Inspired of parallel Private-Shared (PS) MTL and Deep Mutual Learning (DML), we propose KTEPS for MDSC.}
		\label{fig:mtl-fl}
	\end{figure*}
	
	\section{Related Works}
	\subsection{Multi-Domain SC}
	Multi-Domain SC (MDSC) aims to identify sentiment polarity more precisely with the cooperation of multiple domains. Usually, it attempts to distinguish domain shared and domain specific information instead of simply combining all domains' data. \cite{CNN-MTL-ICML2008} shares the word embedding layer among tasks and assigns a specific output branch to each task. \cite{RNN-PS-IJCAI2016} introduces two parallel branches for each individual task, e.g., assigning one separate LSTM layer for each task and meanwhile keeping a shared LSTM layer for all tasks. The illustrations can be found in Figure.~\ref{fig:mtl-fl}. Other advanced MDSC methods take advantage of global shared memory~\cite{ARC-Memory-EMNLP2016}, private attention~\cite{DAEA-IJCAI2019}, or adversarial training~\cite{ASP-MTL-ACL2017,MAN-NAACL-HLT2018} mechanisms for MDSC and have also achieved great success.
	
	\subsection{Federated Learning}
	Federated Learning (FL)~\cite{Fed-Advances,Fed-Concept} gathers participants in a collaborative system, which is tailored for distibuted training with privacy preserved. As categorized in~\cite{Fed-Advances}, cross-silo and cross-device FL mainly differ in amounts of the participants (e.g., $10$ vs. $10^{10}$), communication bottleneck, addressability of clients, etc. The latter one is more relevant to the hardware-level challenges, and we only focus on the cross-silo FL, where small amounts of clients and stable communication can be promised.
	
	Non-IID problem refers to that decentralized samples are typically generated at different contexts, causing challenges to model aggregation and personalization. Various techniques have been proposed to solve the Non-IID problem, e.g., adding regularization term~\cite{FedProx}, sharing a small public dataset~\cite{Fed-Shared-Data,HMR}, taking a fully decentralized paradigm~\cite{Fed-Decentralize-2}, resorting to meta learning~\cite{Fed-MAML}, etc.
	
	\subsection{Aggregation and Personalization}
	In FL, both of the model aggregation and personalization need to be cared. The former is to generate a global model which is applicable to all participants or can be easily adapted to new scenes~\cite{FedAvg,FedProx,Fed-NonIID-Data}, while the latter aims to build invididual qualified models for each participant~\cite{Fed-MultiTask,Fed-Decentralize-2}. 
	In the IID scenario, participants' data are generated from the same distribution, and a single model can achieve both of these two goals at the same time. 
	However, Non-IID data leads to a dilemma that a single aggregated global model can not simultaneously capture the semantics of all participants.
	
	FedAvg~\cite{FedAvg} is purely to aggregate a better global model with the collaboration of multiple participants without consideration of personalization. 
	A direct idea to personalize the global aggregated model is finetuning it on the local data with various settings of hyper-parameters~\cite{FPE}.
	Utilizing fully decentralized learning is natural for better personalization~\cite{Fed-Decentralize-2}, and some meta learning methods have also been investigated~\cite{Fed-MAML}.
	In this paper, we search solutions for considering these two goals simultaneously from the aspect of DNN architecture design, i.e., the private-shared models.
	
	\subsection{Private-Shared Models}
	Private-Shared (PS) models aim to divide private and shared information among domains via feeding data to different network components. \cite{DSN-NeurIPS2016} designs a separation framework for domain adaptation with shared and private encoders. As aforementioned, \cite{CNN-MTL-ICML2008,RNN-PS-IJCAI2016} take single branch and parallel branches for MDSC respectively.
	
	PS models have also been applied to solve Non-IID problem in FL. FedPer~\cite{FedPer-CoRR2019} shares encoder among clients and keeps private classifier for each client for better personalization, expecting the private classifier can capture task-specific knowledge. PFL-DA~\cite{PFL-DA-CoRR2019} keeps an entire model private for each client and shares another model for global aggregation among clients. FURL~\cite{FURL-CoRR2019} keeps the user embedding component private. LG-FedAvg~\cite{LG-FedAvg-CoRR2020} keeps encoder private for each client and shares a global classifier for the heterogeneous multi-modal data. We provide corrsponding correlation analysis between MTL and FL methods when applied to MDSC as illustrated in Figure.~\ref{fig:mtl-fl} and propose KTEPS for FedMDSC.
	
	\subsection{Word Embedding}
	Recent advances represent words as distributed continuous vectors, which can geometrically capture the linguistic regularities and boost the performance of downstream tasks. Although the obtained success, some drawbacks of word embedding have emerged, e.g., implicit bias and huge embedding size. \cite{WV-Gender-Bias-NeurIPS2016,WV-Debias-ACL2020} analyzes the gender bias contained in learned word embeddings and propose methods to remove the sensitive information. \cite{WV-Effective-dimension-reduction} utilizes the post-processing method to reduce embedding dimensions, while~\cite{WV-Distill,WV-Ditill-2} compress word embeddings via distillation. When applied to FL, these problems can lead to acute privacy and transmission problem, and we propose to reduce the dimension of word embeddings via PDR to tackle these challenges.
	
	\begin{figure}
		\includegraphics[width=\linewidth]{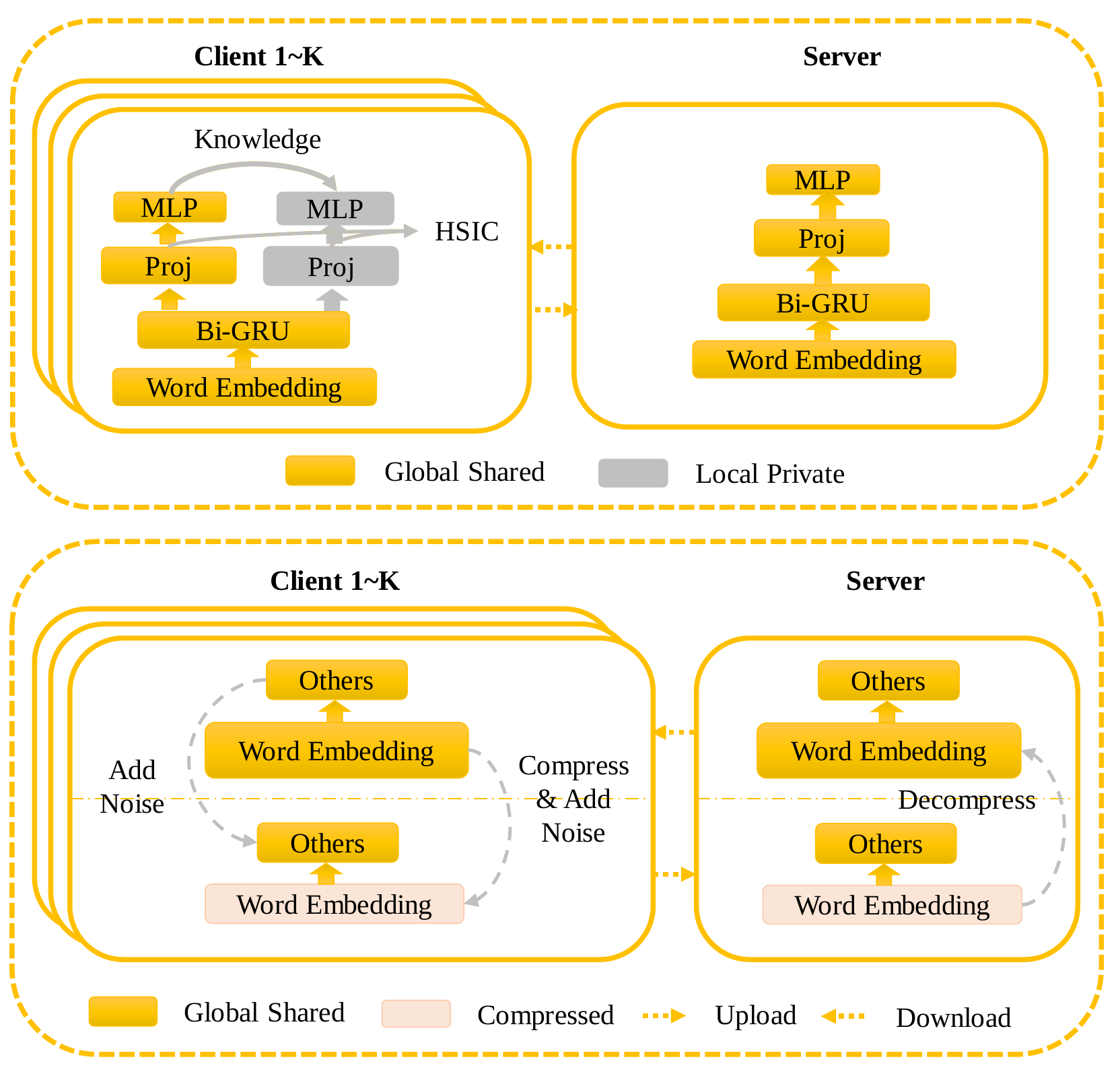}
		\caption{\small Illustration of our proposed methods. The top shows the DNN architecture in the KTEPS for FedMDSC. ``Proj'' refers to a projection layer. The bottom shows the upload and download process with PDR in KTEPS$^\star$.}
		\label{fig:method}
	\end{figure}

	\section{Federated Sentiment Classification}
	\subsection{Basic Settings}
	We limit the scope of our research: {\em we focus on cross-silo federated document-level sentiment classification, and the basic neural network we utilize is a simple ``Embedding-BiRNN-MLP'' architecture as shown in Figure~\ref{fig:mtl-fl}.} Considering other SC paradigms (e.g., aspect-level SC~\cite{AspectSC-AdaRNN-ACL2014,AspectSC-Hier-EMNLP2016}), more complex networks (e.g., transformer~\cite{SC-PivotTransformer-AAAI20}), or cross-device FL (e.g., millions of clients) are future works.
	
	The used simple ``Embedding-BiRNN-MLP'' architecture contains three components, i.e., the word embedding layer, bi-directional rnn (BiRNN) layer and mlp layer, which are abstracted into word embedding (WE), encoder (E), classifier (C) respectively. The word embedding layer maps discrete word indexes to continuous vectors, which are sequentially processed by the BiRNN, and finally the mlp layer predicts the overall sentiment polarity.
	
	\subsection{Basic Notations}
	Suppose we have $K$ participants, and each one has a corpus containing $\Nk$ training reviews and $\Mk$ test reviews. The total number of training samples in the system is $N=\sum_{k=1}^K \Nk$. The reviews are first processed to be sequences of word indexes, and the generated training data and test data of the $k$-th client are $\left\{\left(\bfxk_i, \yk_i\right)\right\}_{i=1}^{\Nk}$ and $\left\{\left(\bfxk_i, \yk_i\right)\right\}_{i=\Nk+1}^{\Nk+\Mk}$ respectively, where $\bfx_i \in \calR^{L}$ is a list of word indexes with a maximum length of $L$ and each $\yk_i \in \{0,1\}^C$ is a one-hot label vector. 
	
	We denote the three components in ``Embedding-BiRNN-MLP'' architecture as three functions, i.e., $E(\cdot)$, $H(\cdot)$, $G(\cdot)$, whose parameters are $\tE, \tH, \tG$ respectively. Specially, the parameters of word embedding layer is denoted as $\tE \in \calR^{V \times d}$, where $V$ is the length of the vocabulary, and $d$ is the dimension of word embeddings. Without more explanation, we use the superscript ``$k$'' for the index of client and subscript ``$i$'' for samples. In MTL or FL for MDSC, symbols without superscript ``$k$'' denote the (global) shared ones. The subscript ``s'' and ``p'' represent ``shared'' and ``private'' respectively.
	
	We denote $\text{Softmax}_{T}\left( \bfg \right)_c = \exp\left( \bfg_{c} / T \right) / \sum_{c^{\prime}} \exp\left( \bfg_{c^{\prime}} / T \right)$ as the softmax function, where $\bfg \in \calR^{C}$ is the predicted ``logits'', the subscript ``$c$'' implies the class index and $T$ is the temperature. Other specific notations will be introduced later.

	\subsection{FL for MDSC} \label{sect:fl-mdsc}
	We first introduce how to apply the most classical FL method (i.e., FedAvg~\citep{FedAvg}) to FedMDSC. For the $k$-th client, it first downloads the global model from the parameter server, i.e., $\tEk \leftarrow \tE$, $\tHk \leftarrow \tH$, $\tGk \leftarrow \tG$, where $\tE, \tH, \tG$ are the global parameters of the three components respectively. Then, it finetunes the model on local data. Take the sample $(\bfxk_i, \yk_i)$ as an example: 
	\begin{eqnarray}
		\bfek_i &=& E\left( \bfxk_i; \tEk \right), \label{eq:embedding} \\
		\bfhk_i &=& H\left( \bfek_i; \tHk \right), \label{eq:rnn} \\
		\bfok_i &=& \text{SeqMean}\left( \bfhk_i \right), \label{eq:seqmean} \\
		\bfgk_i &=& G\left( \bfok_i; \tGk \right), \label{eq:classifier} \\
		\calLk_i &=& -\sum_{c=1}^C \yk_{i,c} \log \left( \text{Softmax}_{1}\left( \bfgk_i \right)_c \right). \label{eq:celoss}
	\end{eqnarray}
	Equation.~\ref{eq:embedding} obtains the word representations $\bfek_i \in \calR^{d \times L}$ through a lookup operation; Equation.~\ref{eq:rnn} utilizes the BiRNN to process the representations and outputs the hidden representations $\bfhk_i \in \calR^{2s \times L}$, where $s$ is the hidden size of the recurrent unit; Equation.~\ref{eq:seqmean} takes the average of the hidden representations along the sequential dimension as input and outputs $\bfok_i \in \calR^{2s}$; Equation.~\ref{eq:classifier} predicts the classification result through a mlp layer and $\bfgk_i \in \calR^{C}$ is the obtained ``logits'' without softmax operator. Finally, we calculate the cross-entropy loss as in Equation.~\ref{eq:celoss}.
	
	Each client finetunes the global model on its local data for some steps and then sends the updated parameters to the parameter server. The server takes a simple model averaging process for these parameters as $\tE = \sum_{k=1}^K \frac{\Nk}{N} \tEk$, $\tH = \sum_{k=1}^K \frac{\Nk}{N} \tHk$, $\tG = \sum_{k=1}^K \frac{\Nk}{N} \tGk$.
	
	The local training procedure and the global aggregation procedure will be repeated several rounds until convergence.
	
	\begin{remark}
		In FedAvg, the data privacy can be protected owing to the fact that only models are transmitted between clients and the parameter server without explicitly sharing users' data. Simultaneously, some other advanced methods (e.g., differential privacy~\cite{DiffPrivate}) can be further applied to obtain stricter privacy protection requirements, which will be discussed later.
	\end{remark}
	
	\subsection{Correlations Between MTL and FL}
	The easiest way to collaborate multiple domains is utilizing a fully-shared network, and training it on all domains' data. Similarly, in FL, FedAvg~\cite{FedAvg} aims to aggregate a single global model, which is shown in Figure.~\ref{fig:mtl-fl} (A).
	Considering PS models in MTL and FL, the single branch MTL~\cite{CNN-MTL-ICML2008} corresponds to the FedPer~\cite{FedPer-CoRR2019}, while the parallel branch MTL~\cite{RNN-PS-IJCAI2016} corresponds to the PFL-DA~\cite{PFL-DA-CoRR2019} as shown in Figure.~\ref{fig:mtl-fl} (B) and (C) respectively. It is notable that FedPer is purely designed for better personalization and it can not generate a complete global model owing to the single branch architecture.
	
	The correlations between MTL and FL when applied to MDSC are obvious. MTL methods divide the whole network into shared and private components according to whether a component serves all tasks or an individual task. {\em Tasks in MTL are equivalent to clients in FL, and the shared components in MTL can be adapted to FL as the globally shared ones which participates in the global model aggregation procedure, while the private ones are kept and updated individually on local clients.}
	Hence, the well-performed MTL methods can be adapted to FL. For example, \cite{DocSC-UPAttn-EMNLP2016} utilizes user and product embedding for better capturing specific information, which can be generalized to privatize the user representation layer on local clients~\cite{FURL-CoRR2019}.
	
	\begin{remark}
		\label{remark:comps}
		Aside from the number of branches, the shared components are also different in Figure.~\ref{fig:mtl-fl} (B) and (C). For example, should we keep a private word embedding layer or BiRNN for each client? We will discuss this in experimental ablation studies.
	\end{remark}
	
	\section{Proposed Methods}
	\subsection{Solution to Non-IID Problem: KTEPS}
	Our goal is to obtain better aggregation and personalization simultaneously even in a Non-IID scene. As aforementioned, FL methods based on single branch networks, e.g., FedPer~\cite{FedPer-CoRR2019} and LG-FedAvg~\cite{LG-FedAvg-CoRR2020}, can not generate a complete model for novel scenes. Hence, we follow FL methods with parallel architectures, e.g., PFL-DA~\cite{PFL-DA-CoRR2019}.
	
	Specifically, we divide the network into several components: global shared word embedding $E_\text{s}(\cdot)$, global shared BiRNN $H_\text{s}(\cdot)$, global shared mlp classifier $G_\text{s}(\cdot)$, local private mlp classifier $G_\text{p}(\cdot)$. The parameters of these components are $\tE_\text{s}$, $\tH_\text{s}$, $\tG_\text{s}$, and $\tG_\text{p}$ respectively. The illustration can be found in the top of Figure.~\ref{fig:method}. Notably, PFL-DA~\cite{PFL-DA-CoRR2019} utilizes a complete private model which may overfit on a small local dataset, while we only privatize a single mlp classification layer, which is introduced in Remark.~\ref{remark:comps} and will be investigated in Section.~\ref{sect:arch}.
	
	Although with the parallel PS architecture, two questions are still naturally asked: {\em (1) How to guarantee that the global shared components can capture client invariant information while the local private ones capture client specific information? Is this PS architecture design enough? (2) How to mitigate the feature mismatch occurring between the newly-downloaded BiRNN and the locally-preserved mlp classification layer?}
	
	\subsubsection{Diversity}
	The first question has been investigated in ASP-MTL~\cite{ASP-MTL-ACL2017}, which proposes to extract domain invariant information with the help of domain adversarial training. However, training a domain discriminator is not an easy task in FL, unless the data features from different domains can be sent out and located in the same device as done in~\cite{FADA-ICLR2020}. We regard that feature sharing will violate the privacy protection constraint. Hence, we relax the requirement of explicitly extracting domain invariant features and only increase the diversity of shared and private classifiers.
	
	Specifically, for the $k$-th client, we project the outputs of the BiRNN, i.e., $o_{i}^k$ calculated as in Equation.~\ref{eq:embedding} to Equation.~\ref{eq:seqmean}, into two different subspaces as shown in top part of Figure.~\ref{fig:method}. Mathematically, the projected features for the $i$-th sample $\bfxk_i$ are obtained as:
	\begin{equation}
		o_{\text{s},i}^k = P_\text{s}\left( o_{i}^k; \tPk_\text{s} \right), \quad
		o_{\text{p},i}^k = P_\text{p}\left( o_{i}^k; \tPk_\text{p} \right), \label{eq:proj}
	\end{equation}
	where $P_\text{s}(\cdot)$ and $P_\text{p}(\cdot)$ are functions of the shared and private projection layers, and $o_{\text{s},i}^k$ and $o_{\text{p},i}^k$ are the projected shared and private features respectively. $\tPk_\text{s}$ and $\tPk_\text{p}$ are parameters of the two projection layers.
	
	To increase the diversity of projected features, we utilize Hilbert-Schmidt Independence Criterion (HSIC)~\cite{HSIC-ALT2005} as a regularization. We adapt the definition of HSIC to our problem as follows:
	\begin{mydef}{(Empirical HSIC)}
		Let $\left\{ \left( o_{\rm{s},i}^k, o_{\rm{p},i}^k \right) \right\}_{i=1}^B$ be a series of $B$ independent observations. An empirical estimation of HSIC is given by:
		\begin{equation}
			\calLk_{\rm{div}} \coloneqq \left(B-1\right)^{-2} \rm{Tr}\left( \bfL_{\rm{s}} \bfH \bfL_{\rm{p}} \bfH \right), \label{eq:divloss}
		\end{equation}
		where $\bfL_{*} \in \calR^{B \times B}$ is the gram matrix defined as $\bfL_{\ast,ij} \coloneqq l_{\ast}\left(o_{\ast,i}^k, o_{\ast,j}^k\right)$, $\ast \in \{\rm{s},\rm{p}\}$. $l_{\ast}$ is a kernel function and $\bfH_{ij}=\calI\{i=j\}-1/B$.
	\end{mydef}
	
	For implementation, we utilize single-layer fully-connected projection mlps with the output's size being the same of the input.

	\subsubsection{Knowledge Transfer}
	For answering the second question, we propose to enhance the information flow from the shared branch to the private one. PFL-DA~\cite{PFL-DA-CoRR2019} only takes a weighted combination strategy to train both branches, which is far from the goal of knowledge transfer. One method to explicitly transfer knowledge is distillation~\cite{KD-CoRR2015}, in which a teacher network can guide the learning process of a student network. Recently, a learning paradigm named Deep Mutual Learning (DML)~\cite{DML-CVPR2018} finds that two networks can co-teach each other simultaneously even they have not yet converged. Inspired of this, we take advantage of DML to enhance the ability of the private classifier.
	
	In detail, we denote the predicted ``logits'' of the two branches as $\bfgk_{\text{s},i} \in \calR^{C}$ and $\bfgk_{\text{p},i} \in \calR^{C}$, and the knowledge transfer loss can be formulated as follows:
	\begin{equation}
		\calLk_{i, \text{kt}} = -\sum_{c=1}^C \text{Softmax}_1\left( \bfgk_{\text{p},i} \right)_c \log \left( \text{Softmax}_T\left( \bfgk_{\text{s},i} \right)_c \right), \label{eq:knowtrans}
	\end{equation}
	where $T$ is the temperature and we stop the gradients of the $\bfgk_{\text{s}}$ in code implementation.
	
	In total, the loss function of a local data batch $\left\{\left(\bfxk_i, \yk_i\right)\right\}_{i=1}^{B}$ is formulated as:
	\begin{equation}
		\calLk_{\text{total}} = \frac{1}{B} \sum_{i=1}^B \left( \calLk_{\text{s},i} + \calLk_{\text{p},i} + \lambda_1 \calLk_{i,\text{kt}}\right) + \lambda_2 \calLk_{\text{div}}, \label{eq:total-loss}
	\end{equation}
	where $\lambda_1$ and $\lambda_2$ are pre-defined coefficients. $\calLk_{\text{s},i}$ and $\calLk_{\text{p},i}$ are cross-entropy losses calculated similarly as in Equation.~\ref{eq:celoss}. The coefficients of these two terms will be investigated in Section.~\ref{sect:loss-terms}.
	
	\subsection{Solution to Word Embedding Problem: PDR and KTEPS$^\star$}
	As aforementioned, word vectors are semantic rich representations which are vulnerable to attacks. For example, as categorized in~\cite{Fed-Advances}, the server can be an honest-but-curious one that can inspect private information through word relationships, e.g., mining the users' jobs or genders through inner product similarities~\cite{WV-Gender-Bias-NeurIPS2016}. Furthermore, the huge word embedding size burdens the transmission a lot. For solving the privacy problem, \cite{PrivateWV-TKDE2018} designs a suite of arithmetic primitives on encrypted word embeddings. However, it is only applicable to the simple CBOW or Skip-gram~\cite{Word2Vec} algorithms and the arithmetic operations are too complex. Hence, we only follow the advocated goals of~\cite{PrivateWV-TKDE2018}, i.e., security, effectiveness, and efficiency. {\em To summarize, the targeted word embeddings should be hard to inspect without degrading the final model's performance a lot, and the computation should be practically acceptable.}
	
	We resort to PDR methods to achieve the above goals simultaneously. Specifically, we utilize PCA to compress word embeddings as done in~\cite{WV-Dimension-PCA,WV-Post-process,WV-Effective-dimension-reduction}. We denote the local word embedding matrix as $\tEk \in \calR^{V \times d}$, and we apply PCA to it:
	\begin{equation}
		\{\bfuk_1, \bfuk_2, \cdots, \bfuk_d\} = \text{PCA}\left( \tEk \right), \label{eq:pca-comps}
	\end{equation}
	where $\{\bfuk_i \in \calR^{d} \}_{i=1}^d$ are principal components sorted by eigenvalues. A normal method is to select the top $d_2$ ones to compose the projection subspace. However, the post-processing method~\cite{WV-Post-process} find that the most of the energy is located in about the top 8 dimensions, and eliminating them can remove the same dominating directions and lead to better performances. Hence, we propose to select the intermediate components $\{\bfuk_{d_1}, \cdots, \bfuk_{d_2}\}$ to form the projection matrix $\bfUk \in \calR^{d \times (d_2 - d_1 + 1)}$, where $1 \leq d_1 < d_2 \leq d$. The compressed word embedding matrix is:
	\begin{equation}
		\hattEk = \tEk \bfUk. \label{eq:wv-compress}
	\end{equation}

	For transmission, we first add noise to both the projection matrix $\bfUk$ and the compressed word embedding matrix $\hattEk$, and then send them to the parameter server. Before aggregation, the server will first decompress $\hattEk$ as follows:
	\begin{equation}
		\ovtEk = \hattEk {\bfUk}^T, \label{eq:wv-decompress}
	\end{equation}
	where we omit the formulation of adding noise.

	\subsubsection{Security} In summary, we add three strategies towards the privacy protection:
	\begin{itemize}
		\item FL is utilized to keep data preserved on local clients without directly transmission.
		\item Noise is added to model parameters to satisfy differential privacy~\cite{DiffPrivate} as much as possible.
		\item Projection based compression can eliminate both dominating directions and subtle information of word embeddings.
	\end{itemize}
	
	We give a detailed analysis for the last one. For a specific word vector $\bfvk$, the server can only restore it as $\ovbfvk = \sum_{i=d_1}^{d_2} \left({\bfvk}^T \bfuk_i\right) \bfuk_i$. On one hand, with larger $d_1$ and smaller $d_2$, it is harder for the server to restore the raw word vectors. On the other hand, the word relationships can be disturbed owing to the fact that:
	\begin{equation}
		\{\ovbfvk_{j_1}\}^{T}\ovbfvk_{j_2} \leq \{\ovbfvk_{j_1}\}^{T}\ovbfvk_{j_3} 
		\nRightarrow 
		\{\bfvk_{j_1}\}^{T}\bfvk_{j_2} \leq \{\bfvk_{j_1}\}^{T}\bfvk_{j_3}, \label{eq:wv-relation}
	\end{equation}
	which can be experimentally proved via the observation that a word pair with larger inner product value can still be larger or become smaller compared with another pair after the embedding compression. The fact implies that the risk of privacy leakage through mining word relationships is decreased, which provides a stricter data privacy protection mechanism.
	
	\subsubsection{Effectiveness} The energy kept is $\sum_{i=d_1}^{d_2}\lambda_{i} / \sum_{i=1}^d \lambda_i$ after the compression, where $\lambda_i$ is the $i$-th singular value of the word embedding matrix. The energy kept and the information removed need to be balanced. In implementations, we find that when $d=200$, setting $d_1=2$ and $d_2=150$ is a good choice. We will investigate the various settings in Section.~\ref{sect:wv-dims}. Additionally, the transmission cost for uploading word embedding has been decreased from $\calO(V \times d)$ to $\calO\left((V + d) \times (d_2 - d_1 + 1)\right)$.
	
	\subsubsection{Efficiency} The PCA algorithm includes computing covariance matrix and applying SVD steps, whose time complexity are $\calO(V\times d^2)$ and $\calO(d^3)$ respectively. When compared to the complex neural network training process, this is efficient enough.
	
	\begin{remark}
		For the $k$-th client, only the vectors of the local vocabulary $\calVk \subset \calV$ can be updated. Hence, we can only download the corresponding word embedding subset from the server and utilize the above PDR method to the subset, further decreasing the uploading transmission cost to $\calO\left((V^k + d) \times (d_2 - d_1 + 1)\right)$ and the computing cost to $\calO(V^k \times d^2 + d^3)$.
	\end{remark}

	Overall, the whole procedure of KTEPS$^\star$ for FedMDSC can be found in Algorithm.~\ref{algo:fed-mdsc}. Notably, KTEPS does not apply the PDR process as in Line.~\ref{line:compress-wv} and Line.~\ref{line:decompress-wv}.	
	 
	\begin{algorithm}
		\caption{KTEPS$^{\star}$} \label{algo:fed-mdsc}
		\begin{algorithmic}[1]
			\small{
				\For{Each Global Round}
				\Procedure{UpdateLocalDevice}{$k$}        \Comment{Update $k$-th local model}
				\State $\tEk_\text{s} \leftarrow \tE_{s}$         \Comment{Download global shared word embedding} 
				\State $\tHk_\text{s} \leftarrow \tH_\text{s}$         \Comment{Download global shared BiRNN} 
				\State $\tPk_\text{s} \leftarrow \tP_\text{s}$         \Comment{Download global shared projection layer} 
				\State $\tGk_\text{s} \leftarrow \tG_\text{s}$         \Comment{Download global shared mlp classifier}
				\For{Each Local Epoch}
					\For{Each Data Batch $\{(\bfxk_{i},\yk_{i})\}_{i=1}^{B}$}
					\State Calculate loss as in Equation.~\ref{eq:total-loss}
					\State Update $\tEk_\text{s},\tHk_\text{s},\tPk_\text{s},\tGk_\text{s}$ and $\tPk_\text{p},\tGk_\text{p}$
					\EndFor
				\EndFor
				\State Compress $\tEk_\text{s}$ as in Equation.~\ref{eq:pca-comps} and ~\ref{eq:wv-compress}, get $\hat{\theta}^k_{E_\text{s}}$ and $\bfUk$
				\label{line:compress-wv}
				\State Add noise to $\hat{\theta}^k_{E_\text{s}},\bfUk,\tHk_\text{s},\tPk_\text{s},\tGk_\text{s}$ and send them to server
				\EndProcedure
				
				\Procedure{UpdateGlobalServer}{}        \Comment{Update global model}
				\State Decompress $\hat{\theta}^k_{E_\text{s}}$ as in Equation.~\ref{eq:wv-decompress}, get $\ovtEk$
				\label{line:decompress-wv}
				\State $\tE_s = \sum_{k=1}^K \frac{\Nk}{N} \ovtEk_s$,  $\tH_s = \sum_{k=1}^K \frac{\Nk}{N} \tHk_s$, $\tP_s = \sum_{k=1}^K \frac{\Nk}{N} \tPk_s$, $\tG_s = \sum_{k=1}^K \frac{\Nk}{N} \tGk_s$    \Comment{Parameter aggregation}
				\EndProcedure
				\EndFor
			}
		\end{algorithmic}
	\end{algorithm}
	
	\section{Experiments}
	\subsection{Datasets and Preprocessing Details} \label{sect:preprocess}
	We construct two FedMDSC scenes, i.e., \textbf{FDU} and \textbf{IYY}. FDU\footnote{\url{http://nlp.fudan.edu.cn/data/}} contains 16 SC tasks, which is originally proposed to verify the ASP. method~\cite{ASP-MTL-ACL2017}. We distribute the 16 tasks onto 16 clients. Each client in FDU does a 2-class SC task. IYY contains only 3 clients, which is constructed by three common SC benchmarks, i.e., IMDB, Yelp13, Yelp14\footnote{\url{http://ir.hit.edu.cn/~dytang/}}. Both Yelp13 and Yelp14 contain 5 sentiment levels, while IMDB has 10 levels. For consistency, we discretize the 10 sentiment levels in IMDB into 5 levels by merging two successive levels. In FDU and IYY, each client has a local train and test set. We do not use validation sets, and the used evaluation criterions will be introduced in Section.~\ref{sect:metric}.
	
	We utilize the word segmentations provided in the downloaded corpus and only split reviews via space. We count the word occurrences for each client individually and then aggregate them on the parameter server. We select the most 50000 frequent ones as the global vocab $\calV$ for both FDU and IYY. We clip or pad the reviews to have the maximum length 200 for FDU and 400 for IYY. Additionally, we add ``<unk>'' to denote the words that are not in the vocab and ``<pad>'' to denote the padded words. The statistical information of the two scenes can be found in Table.~\ref{tab:fdu-iyy-info}.
	
	\begin{table}
		\centering
		\caption{\small The statistical information of the FDU and IYY scene.} \label{tab:fdu-iyy-info}
		{
			\begin{tabular}{ccccccc}
				\hline \hline
				Scene & Client & $\Nk$ & $\Mk$ & Avg.$L$ & $ V^k $ & $C$ \\
				\hline \hline
				\multirow{16}{*}{\textbf{FDU}}
				& Apparel & 1600 & 400 & 71 & 7K & 2 \\
				& Baby & 1500 & 400 & 125 & 8K & 2 \\
				& Books & 1600 & 400 & 190 & 22K & 2 \\
				& Camera & 1597 & 400 & 146 & 10K & 2 \\
				& DVD & 1600 & 400 & 208 & 21K & 2 \\
				& Elec. & 1598 & 400 & 124 & 9K & 2 \\
				& Health & 1600 & 400 & 99 & 8K & 2 \\
				& Kitchen & 1600 & 400 & 104 & 8K & 2 \\
				& Mag. & 1570 & 400 & 135 & 11K & 2 \\
				& Music & 1600 & 400 & 157 & 14K & 2 \\
				& Soft. & 1515 & 400 & 156 & 9K & 2 \\
				& Sports & 1599 & 400 & 114 & 9K & 2 \\
				& Toys & 1600 & 400 & 108 & 9K & 2 \\
				& Video & 1600 & 400 & 173 & 17K & 2 \\
				& Imdb & 1600 & 400 & 312 & 21K & 2 \\
				& MR & 1600 & 400 & 23 & 7K & 2 \\
				\hline \hline
				\multirow{3}{*}{\textbf{IYY}}
				& Imdb & 67426 & 9112 & 409 & 44K & 5 \\
				& Yelp13 & 62522 & 8671 & 199 & 38K & 5 \\
				& Yelp14 & 183019 & 25399 & 207 & 43K & 5 \\
				\hline \hline
			\end{tabular}
		}
	\end{table}
	
	\subsection{Network and Hyper-parameters}
	\label{sect:hyperparam}
	We utilize the aforementioned ``Embedding-BiRNN-MLP'' as the base model. Specifically, we set the word embedding size as 200 and initialize it with Glove\footnote{\url{http://nlp.stanford.edu/data/glove.6B.zip}}. We utilize BiGRU with a hidden size of 64 as BiRNN, and a two layer fully connected network with ReLU activation for classification.
	
	We use SGD with a momentum of 0.9 as the optimizer. We set LR=0.01 for FDU and set LR=0.1 for IYY. Additionally, we find that setting the LR of the embedding layer in IYY to be 0.01 can be better. We set the batch size of FDU and IYY to be 8 and 64 correspondingly. For FL methods, we set the number of local epoches to be 2 and the global aggregation round is iterated 50 and 20 times for FDU and IYY respectively. We add gaussian noise with $\sigma=0.01$ to the model parameters independently. For KTEPS, we utilize gaussian kernel function in Equation.~\ref{eq:divloss}, $T=0.25$ in Equation.~\ref{eq:knowtrans} and set $\lambda_1=0.01, \lambda_2=0.01$ in Equation.~\ref{eq:total-loss}. For KTEPS$^\star$, we set $d_1=2$ to remove the largest principal component of word embeddings and $d_2=150$ to reduce the communication cost. For other hyper-parameters in compared methods, we utilize the settings reported in corresponding papers.
		
	\subsection{Evaluation Metric}
	\label{sect:metric}
	For evaluation of global model aggregation, we calculate the global model's accuracy on all participants' test data:
	\begin{equation}
		\text{Ag}^k = \frac{1}{\Mk}\sum_{i=\Nk+1}^{\Nk+\Mk}\calI\left\{\arg\max \bfgk_i = \arg\max \yk_i \right\},
		\label{eq:acc}
	\end{equation}
	\begin{equation}
		\text{Ag} = \sum_{k=1}^K \frac{1}{K} \text{Ag}^k,
		\label{eq:agg-eval}
	\end{equation}
	where $\calI\{\cdot\}$ is the indication function. $\text{Ag}^k$ is the accuracy for the $k$-th client and $\text{Ag}$ is the averaged result.
	
	For evaluation of model personalization, we take a different but similar metric as in~\cite{FPE}. Upon the aggregation stage converges, we finetune the global model on local clients' training data for $Z$ steps and test the model on the local test data every $z$ steps. We set $Z = n_Z \times z$, and a group of test accuracies can be recorded as $\left\{\text{Ap}_{t}^k\right\}_{t=1}^{n_Z}$. We denote the personalization ability as:
	\begin{equation}
		\text{Ap}^k = \sum_{t=1}^{n_Z} \frac{1}{n_Z} \text{Ap}_{t}^k, \quad \text{Ap} = \sum_{k=1}^K \frac{1}{K} \text{Ap}^k,
		\label{eq:per-eval}
	\end{equation}
	
	\begin{remark}
		In FedAvg, the $\text{Ag}^k$ is a special case of $\text{Ap}_{t}^k$ when $t=0$, i.e., no any finetune steps. For $t>0$ or other PS frameworks, $\text{Ag}^k$ is evaluated on global aggregated model, while $\text{Ap}^k$ is evaluated on both shared and private components. For KTEPS and KTEPS$^\star$, we average the outputs of private and shared classifiers for calculating $\text{Ap}^k$, and different inference mechanisms are investigated in Section.~\ref{sect:local-infer}.
	\end{remark}

	\begin{table*}
		\centering
		\caption{\small Model aggregation results of our models on FDU and IYY against typical baselines. We list $\text{Ag}^k$ for all clients and report the $\text{Ag}$ in the ``Avg'' row filled in gray.} \label{tab:agg-results}
		{
			\begin{tabular}{c|c|ccc|ccccc|cc}
				\hline \hline
				Scene & Client & Indiv. & FS & ASP. & FedAvg & FedFu. & PFL. & FedProx & FedMMD & KTEPS & KTEPS$^{\star}$ \\
				\hline \hline
				\multirow{16}{*}{\textbf{FDU}}
				& Apparel & 83.6 & 86.4 & 86.0 & 86.2 & 84.1 & 85.2 & 86.0 & 87.6 & 85.7 & 84.4\\
				& Baby & 81.1 & 89.0 & 88.5 & 85.2 & 85.4 & 88.4 & 87.8 & 88.1 & 89.8 & 85.2\\
				& Books & 79.1 & 82.5 & 85.4 & 83.3 & 82.6 & 84.6 & 81.8 & 84.1 & 82.7 & 80.2\\
				& Camera & 84.9 & 85.3 & 88.8 & 86.2 & 82.4 & 85.4 & 86.6 & 85.2 & 86.6 & 83.4\\
				& DVD & 80.7 & 84.5 & 86.2 & 87.5 & 85.8 & 85.4 & 83.6 & 84.9 & 86.3 & 81.0\\
				& Elec. & 80.1 & 85.2 & 85.4 & 86.1 & 84.1 & 84.1 & 84.0 & 84.9 & 84.2 & 85.2\\
				& Health & 79.8 & 89.0 & 90.2 & 87.6 & 86.0 & 88.0 & 87.9 & 88.5 & 89.2 & 86.5\\
				& Kitchen & 79.8 & 86.8 & 87.9 & 86.1 & 86.8 & 88.2 & 87.2 & 88.8 & 87.7 & 86.9\\
				& Mag. & 89.6 & 87.7 & 87.5 & 87.8 & 85.6 & 87.6 & 84.8 & 86.1 & 85.7 & 85.6\\
				& Music & 76.6 & 82.7 & 82.2 & 85.7 & 79.8 & 81.8 & 80.8 & 80.4 & 83.3 & 79.9\\
				& Soft. & 85.0 & 89.4 & 87.4 & 86.6 & 85.9 & 86.8 & 87.2 & 88.0 & 88.9 & 85.1\\
				& Sports & 82.1 & 84.3 & 86.9 & 84.9 & 84.5 & 85.4 & 86.4 & 84.0 & 85.6 & 83.5\\
				& Toys & 83.3 & 89.9 & 89.6 & 86.2 & 86.2 & 88.8 & 86.6 & 88.9 & 88.2 & 90.6\\
				& Video & 82.8 & 88.1 & 86.2 & 87.1 & 85.9 & 87.2 & 85.4 & 86.0 & 87.6 & 83.5\\
				& Imdb & 77.2 & 83.1 & 82.9 & 85.3 & 82.8 & 82.8 & 82.5 & 83.5 & 84.0 & 79.4\\
				& MR & 72.4 & 75.8 & 74.4 & 69.2 & 73.4 & 74.8 & 74.4 & 74.8 & 75.3 & 65.9\\
				
				\hline
				\rowcolor{gray!40}
				& Avg & 81.1 & 85.6 & 86.0 & 85.1 & 83.8 & 85.3 & 84.6 & 85.2 & \bf{85.7} & 82.9\\
				\hline \hline
				\multirow{3}{*}{\textbf{IYY}}
				& Imdb & 55.3 & 55.7 & 56.7 & 54.6 & 53.7 & 53.6 & 56.6 & 53.4 & 55.6 & 54.9\\
				& Yelp13 & 61.8 & 62.6 & 63.8 & 62.4 & 61.6 & 61.5 & 61.0 & 59.4 & 62.9 & 62.3\\
				& Yelp14 & 59.7 & 60.1 & 61.0 & 60.1 & 59.3 & 59.4 & 59.3 & 57.1 & 60.7 & 60.0\\
				
				\hline
				\rowcolor{gray!40}
				& Avg & 58.9 & 59.5 & 60.5 & 59.1 & 58.2 & 58.2 & 59.0 & 56.6 & \bf{59.7} & 59.1\\
				\hline \hline
			\end{tabular}
		}
	\end{table*}
	
	\begin{table*}
		\centering
		\caption{\small Model personalization results of our models on FDU and IYY against typical baselines. We list $\text{Ap}^k$ for all clients and report the $\text{Ap}$ in the ``Avg'' row filled in gray.} \label{tab:per-results}
		{
			\begin{tabular}{c|c|ccc|ccccccc|cc}
				\hline \hline
				Scene & Client & Indiv. & FS & ASP. & FedAvg & FedPer & LG. & FedFu. & PFL. & FedProx & FedMMD & KTEPS & KTEPS$^{\star}$ \\
				\hline \hline
				\multirow{16}{*}{\textbf{FDU}}
				& Apparel & 83.6 & 86.8 & 86.0 & 86.6 & 87.2 & 80.4 & 84.2 & 86.4 & 86.0 & 87.4 & 86.2 & 85.0\\
				& Baby & 81.1 & 89.1 & 88.0 & 86.5 & 87.6 & 82.0 & 85.3 & 83.9 & 87.1 & 88.0 & 89.8 & 84.9\\
				& Books & 79.1 & 82.7 & 84.7 & 84.7 & 81.9 & 81.0 & 82.2 & 80.7 & 84.0 & 84.2 & 83.6 & 80.5\\
				& Camera & 84.9 & 85.5 & 88.2 & 85.8 & 86.3 & 85.2 & 85.7 & 85.2 & 86.2 & 85.0 & 86.8 & 82.5\\
				& DVD & 80.7 & 84.5 & 86.2 & 85.0 & 83.4 & 78.3 & 81.5 & 81.4 & 83.8 & 84.9 & 86.0 & 80.8\\
				& Elec. & 80.1 & 85.0 & 86.3 & 86.2 & 84.3 & 80.5 & 82.3 & 79.8 & 85.6 & 84.9 & 84.0 & 84.0\\
				& Health & 79.8 & 88.6 & 90.5 & 89.8 & 88.0 & 79.6 & 84.5 & 84.3 & 87.8 & 88.4 & 89.8 & 87.6\\
				& Kitchen & 79.8 & 86.4 & 87.4 & 84.0 & 88.9 & 80.8 & 79.8 & 80.0 & 87.1 & 88.5 & 87.3 & 86.0\\
				& Mag. & 89.6 & 87.6 & 87.4 & 87.5 & 87.8 & 91.2 & 88.5 & 89.5 & 85.3 & 87.9 & 87.5 & 86.9\\
				& Music & 76.6 & 82.8 & 83.1 & 84.0 & 80.7 & 76.9 & 79.1 & 79.0 & 82.1 & 82.1 & 82.2 & 79.0\\
				& Soft. & 85.0 & 90.0 & 89.0 & 87.2 & 88.6 & 88.5 & 85.0 & 81.5 & 88.5 & 88.2 & 88.9 & 85.5\\
				& Sports & 82.1 & 84.6 & 85.4 & 84.2 & 85.7 & 81.5 & 84.8 & 83.3 & 84.9 & 84.0 & 85.3 & 83.1\\
				& Toys & 83.3 & 89.8 & 90.2 & 88.2 & 88.0 & 81.8 & 83.3 & 86.2 & 88.0 & 89.0 & 89.0 & 91.2\\
				& Video & 82.8 & 88.0 & 86.6 & 86.7 & 87.3 & 84.2 & 86.4 & 81.5 & 85.7 & 86.3 & 87.6 & 84.4\\
				& Imdb & 77.2 & 82.7 & 82.6 & 81.8 & 82.5 & 78.0 & 76.5 & 78.0 & 82.0 & 83.9 & 82.8 & 79.2\\
				& MR & 72.4 & 76.0 & 75.2 & 77.0 & 75.6 & 69.3 & 70.9 & 72.4 & 74.9 & 76.2 & 75.7 & 69.7\\
				
				\hline
				\rowcolor{gray!40}
				& Avg & 81.1 & 85.6 & 86.0 & 85.3 & 85.2 & 81.2 & 82.5 & 82.1 & 84.9 & 85.6 & \bf 85.8 & 83.1\\
				\hline \hline
				\multirow{3}{*}{\textbf{IYY}}
				& Imdb & 55.3 & 56.5 & 56.2 & 56.0 & 56.0 & 60.0 & 57.8 & 57.8 & 57.6 & 53.7 & 56.9 & 55.4\\
				& Yelp13 & 61.8 & 63.1 & 64.1 & 62.6 & 62.4 & 62.0 & 62.0 & 61.6 & 61.4 & 59.7 & 63.0 & 62.8\\
				& Yelp14 & 59.7 & 60.2 & 61.4 & 60.3 & 60.1 & 60.5 & 60.1 & 59.8 & 59.5 & 57.7 & 60.9 & 60.4\\
				\hline
				\rowcolor{gray!40}
				& Avg & 58.9 & 59.9 & 60.6 & 59.6 & 59.5 & \bf 60.8 & 59.9 & 59.7 & 59.5 & 57.0 & 60.3 & 59.6\\
				\hline \hline
			\end{tabular}
		}
	\end{table*}

	\subsection{Compared Methods}
	We briefly introduce the compared methods as follows:
	\begin{itemize}
		\item \textbf{Indiv.} trains model on each client individually.
		\item \textbf{FS} trains model in a fully shared manner without data privacy protection (Figure.~\ref{fig:mtl-fl} (A)).
		\item \textbf{ASP.}~\cite{ASP-MTL-ACL2017} utilizes adversarial MTL to separate domain shared and domain specific information without data privacy protection.
		\item \textbf{FedAvg}~\cite{FedAvg} trains a single model for all clients with data privacy protection (Figure.~\ref{fig:mtl-fl} (A)).
		\item \textbf{FedPer}~\cite{FedPer-CoRR2019} takes single branch for each client and keeps only the mlp classifier private (Figure.~\ref{fig:mtl-fl} (B)).
		\item \textbf{FedFusion}~\cite{FedFusion} fuses the local and global model's features to accelerate FL.
		\item \textbf{LG.}~\cite{LG-FedAvg-CoRR2020} keeps the embedding layer and the BiRNN private, while shares the classifier globally.
		\item \textbf{PFL.}~\cite{PFL-DA-CoRR2019} keeps a complete model shared and trains a private model for each client individually (Figure.~\ref{fig:mtl-fl} (C)).
		\item \textbf{FedProx}~\cite{FedProx} adds a proximal term to regularize the local update of FedAvg to solve the Non-IID problem.
		\item \textbf{FedMMD}~\cite{FedMMD} utilizes a MMD based regularization term, making the features of local model similar to the global model.
		\item \textbf{KTEPS} is the proposed method (Figure.~\ref{fig:method} top). We add the diversity (Equation.~\ref{eq:divloss}) and knowledge transfer (Equation.~\ref{eq:knowtrans}) terms to enhance both of the model aggregation and personalization performances.
		\item \textbf{KTEPS}$^{\star}$ utilizes projection based word embedding compression to KTEPS (Figure.~\ref{fig:method} bottom).
	\end{itemize}
	
	Notably, FedPer and LG. can not aggregate a complete global model owing to the single branch architecture, so we do not compare our methods against them in global model aggregation.

	\subsection{Experimental Results}
	The model aggregation and personalization results are listed in Table.~\ref{tab:agg-results} and Table.~\ref{tab:per-results}. We list $\text{Ag}^k$ and $\text{Ap}^k$ for all clients and report $\text{Ag}$ and $\text{Ap}$ in the ``Avg'' row. Indiv. displays the performance lowerbound, while FS simply combines data together and is not a proper upperbound. ASP. obtains best results owing to explicit domain adversarial training. KTEPS can almost get higher performances than FedAvg and FS, while a bit weaker than ASP. as expected. Furthermore, KTEPS can get better results than the compared FL variants. Compared with KTEPS, the performances of KTEPS$^\star$ drop a lot, whereas it can still work better than individual training.
	
	From the aspect of clients, we find that some clients will get a performance degradation in MTL or FL. For example, the Mag. in FDU can reach 89.6 in individual training, whereas it can only reach 87.5 and 85.7 in ASP. and KTEPS. Another observation is that Yelp13 and Yelp14 can get higher gains in IYY, while Imdb gets little improvement. These observations are related to task similarities. Mining correlated tasks or detecting outliers are future works.
	
\begin{figure}
		\includegraphics[width=\linewidth]{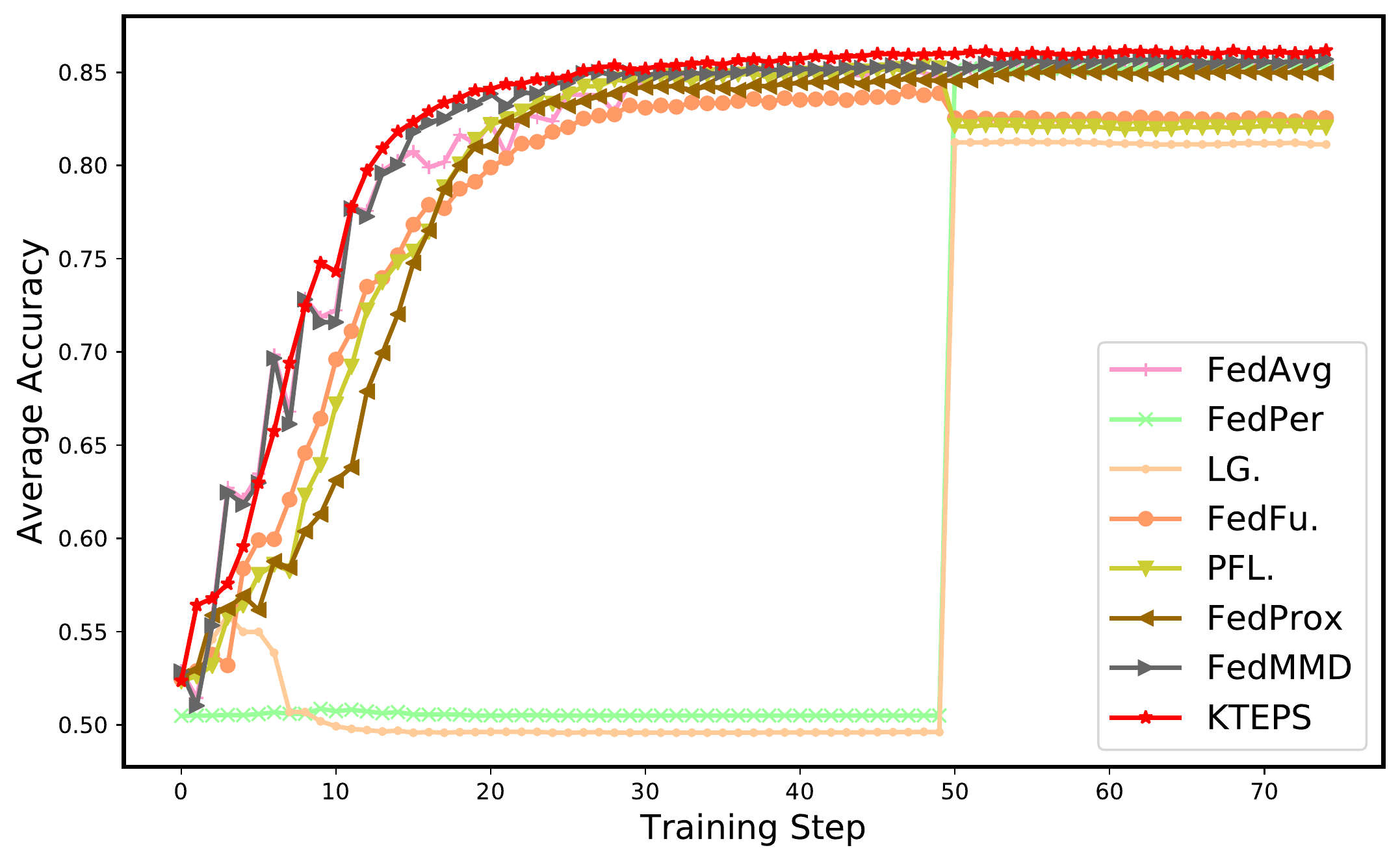}
		\caption{\small Convergence curves of FL methods on FDU. The first 50 steps display the $\text{Ag}$ results and the left 25 steps display $\text{Ap}_t$.}
		\label{fig:converge-fdu}
	\end{figure}

	Additionally, to intuitively display the superiorities of KTEPS, we record the $\text{Ag}$ in each aggregation round and $\text{Ap}_t$ for each $t$-th personalization step. We plot $\text{Ag}$ and $\text{Ap}_t$ of FDU in Figure.~\ref{fig:converge-fdu}. First, FerPer and LG. obtains $\text{Ag} \approx 0.5$ owing to that they can not obtain complete global models, while the personalization results are normal. Second, the personalization performances of LG., FedFu., and PFL. are lower than other methods because these three methods both keep a word embedding layer private, which causes overfitting. FedMMD can weakly obtain superior results than FedAvg and FedProx, and KTEPS can get best aggregation and personalization results simultaneously.
	
	\subsection{Ablation Studies}
	\subsubsection{Loss Terms and Coefficients}
	\label{sect:loss-terms}
	We first study the loss terms in Equation.~\ref{eq:total-loss}. We set $\lambda_1 \in \{0.0, 0.001, 0.01\}, \lambda_2 \in \{0.0, 0.001, 0.01\}$ and report $\text{Ag}$ and $\text{Ap}$ in Table.~\ref{tab:ablation-loss-coef}. Specially, setting $\lambda_1=0.0$ and $\lambda_2=0.0$ is similar to FedAvg, while the network architecture is different. We can find that with $\lambda_1 > 0.0$ or $\lambda_2 > 0.0$, the performances become better. We also use larger $\lambda_2=\{0.1, 1.0\}$, and the performances drop a lot. We analyze the logged loss values and observe that the diversity loss have a higher magnitude. Hence, empirically setting $\lambda_2 \in [0.001, 0.01]$ will be an appropriate choice.
	
	\begin{table}
		\centering
		\caption{\small Model aggregation and personalization results of our models on FDU and IYY with different loss coefficients.} \label{tab:ablation-loss-coef}
		{
			\begin{tabular}{c|c|c|c|c|c|c|c}
				\hline \hline
				& & \multicolumn{3}{c|}{$\text{Ag}$} & \multicolumn{3}{c}{$\text{Ap}$} \\
				\hline
				Scene & \diagbox{$\lambda_1$}{$\lambda_2$} & 0.0 & 0.001 & 0.01 & 0.0 & 0.001 & 0.01 \\
				\hline \hline
				\multirow{3}{*}{\textbf{FDU}}
				& 0.0 & 85.0 & 85.4 & 85.2 & 85.2 & 85.5 & 85.3\\
				& 0.001 & 85.5 & 85.3 & 85.1 & 85.8 & 85.6 & 85.3\\
				& 0.01 & 85.2 & 85.0 & 85.7 & 85.3 & 85.3 & 85.8\\
				
				\hline \hline
				\multirow{3}{*}{\textbf{IYY}}
				& 0.0 & 58.8 & 59.0 & 58.8 & 59.6 & 59.6 & 59.7\\
				& 0.001 & 59.1 & 59.2 & 59.1 & 59.7 & 59.7 & 59.8\\
				& 0.01 & 58.5 & 59.5 & 59.7 & 59.7 & 59.7 & 60.3\\
				\hline \hline
			\end{tabular}
		}
	\end{table}

	\subsubsection{Personalization Mechanisms}
	\label{sect:local-infer}
	For model personalization, setting a smaller LR is very important. We set the personalization LR as $\mu * \text{LR}_{\text{ag},0}$, where $\text{LR}_{\text{ag},0}$ is the initial LR used in global aggregation stage as aforementioned in Section.~\ref{sect:hyperparam}. The $\text{Ap}$ results are listed in Table.~\ref{tab:ablation-per-lr}. If we set $\mu=1.0$, it is too large and the performances drop a lot. A smaller $\mu \in [0.001, 0.01]$ can lead to better personalization performances.
	
	\begin{table}
		\centering
		\caption{\small Model personalization results of our models on FDU and IYY with different settings of learning rate multiplier $\mu$.} \label{tab:ablation-per-lr}
		{
			\begin{tabular}{c|c|c|c|c|c|c|c|c}
				\hline \hline
				& \multicolumn{4}{c|}{KTEPS} & \multicolumn{4}{c}{KTEPS$^{\star}$} \\
				\hline
				\diagbox{S.}{$\mu$} & 1.0 & 0.1 & 0.01 & 0.001 & 1.0 & 0.1 & 0.01 & 0.001 \\
				\hline \hline
				\textbf{FDU} 
				& 79.0 & 85.1 & 85.8 & 85.5 & 77.5 & 81.1 & 83.1 & 83.5\\
				\hline \hline
				\textbf{IYY}
				& 44.5 & 58.1 & 60.3 & 59.9 & 40.3 & 59.1 & 59.6 & 59.4\\
				\hline \hline
			\end{tabular}
		}
	\end{table}

	In addition, we find that our proposed methods are invariant to local inference mechanisms, i.e., making predictions via the shared classifier (``s''), private classifier (``p''), or the average of them (``sp''). We report $\text{Ap}$ with these three inference ways (``IW'') as in Table.~\ref{tab:ablation-per-local-infer}. Although ``p'' and ``sp'' are weakly better than ``s'', the results are almost the same. First, the added knowledge transfer term makes the predictions of private and shared classifier consistent. Second, the private classifier captures domain specific information and can get weakly better personalization results.

	\begin{table}
		\centering
		\caption{\small Model personalization results of our models on FDU and IYY with different local inference mechanisms.} \label{tab:ablation-per-local-infer}
		{
			\begin{tabular}{c|c|c|c|c|c|c}
				\hline \hline
				 & \multicolumn{3}{c|}{KTEPS} & \multicolumn{3}{c}{KTEPS$^{\star}$} \\
				\hline
				\diagbox{S.}{IW} & s & p & sp & s & p & sp \\
				\hline \hline
				\textbf{FDU}
				& 85.5 & 85.6 & 85.8 & 83.0 & 83.0 & 83.1\\
				\hline \hline
				\textbf{IYY}
				& 60.1 & 60.2 & 60.3 & 59.5 & 59.6 & 59.6 \\
				\hline \hline
			\end{tabular}
		}
	\end{table}
	
	\begin{table}
		\centering
		\caption{\small Model aggregation and personalization results of our models on FDU and IYY with different $d_1$ and $d_2$ settings.} \label{tab:ablation-comp-dims}
		{
			\begin{tabular}{c|c|c|c|c|c|c|c|c|c}
				\hline \hline
				& & \multicolumn{4}{c|}{$\text{Ag}$} & \multicolumn{4}{c}{$\text{Ap}$} \\
				\hline
				S. & \diagbox{$d_2$}{$d_1$} & 1 & 2 & 3 & 6 & 1 & 2 & 3 & 6 \\
				\hline \hline
				\multirow{2}{*}{\textbf{F.}}
				& 200 & 85.7 & 83.7 & 82.6 & 50.5 & 85.8 & 83.9 & 83.1 & 49.2\\
				& 150 & 84.9 & 82.9 & 61.6 & 50.5 & 85.5 & 83.1 & 65.0 & 49.1\\
				\hline \hline
				\multirow{2}{*}{\textbf{I.}}
				& 200 & 59.7 & 59.0 & 58.7 & 58.2 & 60.3 & 59.9 & 59.5 & 59.2\\
				& 150 & 59.5 & 59.1 & 58.8 & 56.9 & 60.3 & 59.6 & 59.0 & 58.1\\
				\hline \hline
			\end{tabular}
		}
	\end{table}

	\begin{table}
		\centering
		\caption{\small Model aggregation and personalization results of our models on FDU with different architectures and different $\lambda_1$.} \label{tab:ablation-arch}
		{
			\begin{tabular}{c|c|c|c|c|c|c}
				\hline \hline
				 & \multicolumn{3}{c|}{$\text{Ag}$} & \multicolumn{3}{c}{$\text{Ap}$} \\
				\hline
				\diagbox{$\lambda_1$}{Arch.} & A & B & C & A & B & C \\
				\hline \hline
				0.0 & 85.0 & 85.1 & 85.3 & 85.6 & 83.5 & 82.1\\
				0.001 & 85.6 & 85.2 & 85.1 & 86.0 & 83.6 & 82.8\\
				0.01 & 85.8 & 85.5 & 85.3 & 85.8 & 83.3 & 81.4\\
				0.1 & 85.9 & 85.5 & 85.6 & 86.0 & 83.7 & 82.1\\
				\hline \hline
			\end{tabular}
		}
	\end{table}

	\subsubsection{Word Embedding Compression}
	\label{sect:wv-dims}
	The settings of $d_1$ and $d_2$ in KTEPS$^\star$ are vital to the performances. We set $d_1 \in \{1, 2, 3, 6\}$ and $d_2 \in \{200, 150\}$, where $d_1=1$, $d_2=200$ degenerates into KTEPS. The $\text{Ag}$ and $\text{Ap}$ results are listed in Table.~\ref{tab:ablation-comp-dims}. The performances drop sharply with $d_1$ becoming larger. This is reasonable because removing the top principal components leads to information loss. However, this is inconsistent with PPA~\cite{WV-Post-process}, which may owing to that PPA is a post-processing method while KTEPS$^{\star}$ takes an iterated procedure which can raise error accumulation. All in all, we can remove the first component and the subtle 50 dimensions, e.g., $d_1=2$ and $d_2=150$, for protecting privacy and reducing transmission cost respectively, and the performances are acceptable. Notably, if we assume the privacy constraint on word embeddings can be relaxed and we only remove the smallest 50 dimensions for releasing the transmission burden, e.g., $d_1=1$ and $d_2=150$, the performances can achieve much better results.

	\begin{figure}
		\includegraphics[width=\linewidth]{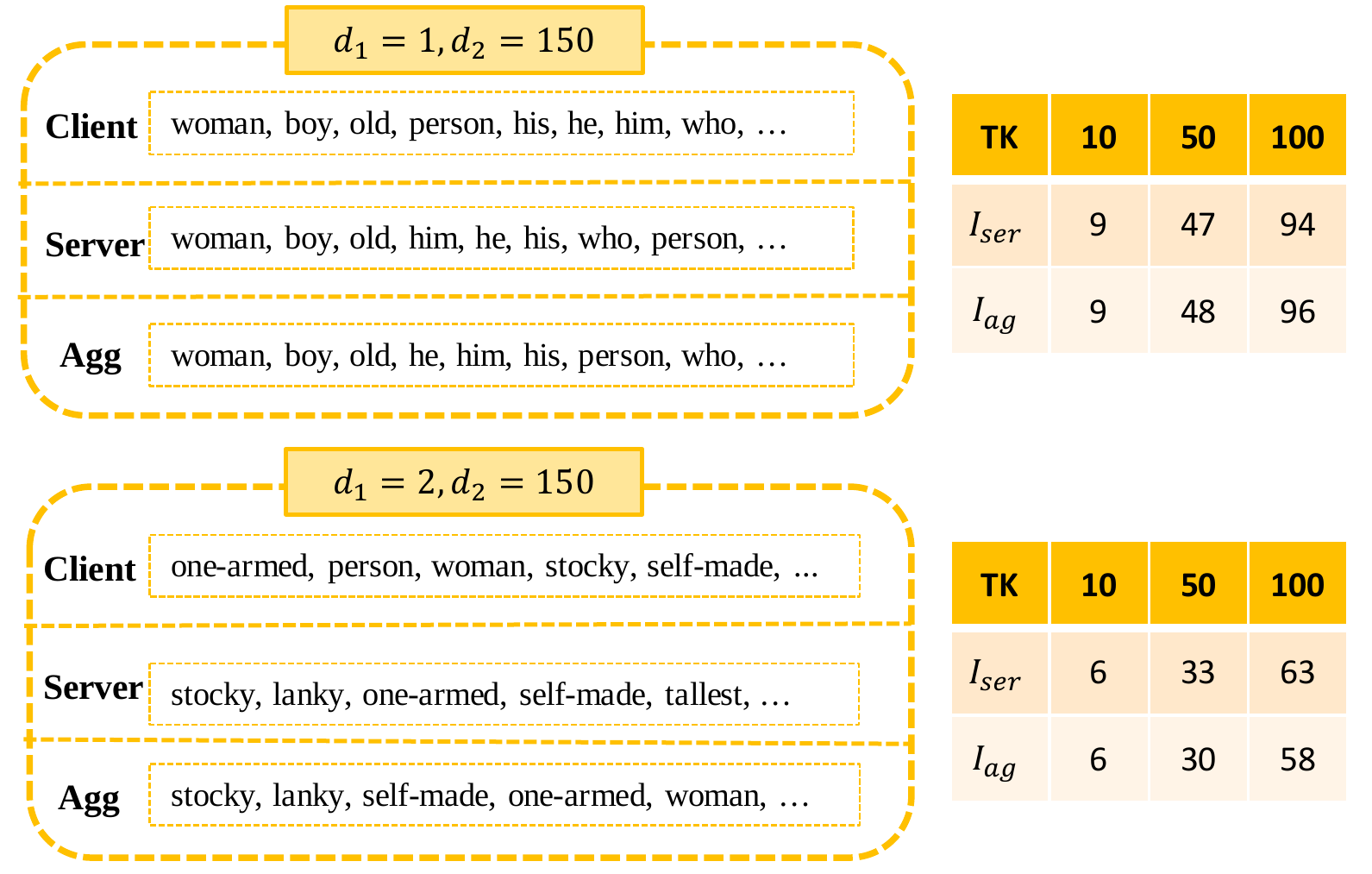}
		\caption{\small Illustration of the privacy protection mechanism in IYY.}
		\label{fig:wv-privacy}
	\end{figure}

	We present an empirical explanation for the privacy protection mechanism of KTEPS$^\star$ in IYY. We compare the most similar words of ``man'' in three word embeddings, i.e., the one on client ``Yelp13'' before compression, the corresponding one that the server restores (Equation.~\ref{eq:wv-decompress}), the final aggregated one which will be sended to clients. We denote these as ``Client'', ``Server'', and ``Agg'' in Figure.~\ref{fig:wv-privacy}. We first qualitatively present the most similar words of ``man'' in the dashed boxes, where we find that setting $d_1=1,d_2=150$ can not disturb the word relationships a lot, while setting $d_1=2$ can lead to distinct similar words. For quantitatively analysis, we calculate $\bf{I}_{\text{ser}}$ as the number of intersected top-$K$ similar words in ``Server'' and ``Client'', and $\bf{I}_{\text{ag}}$ for ``Agg'' and ``Client''. A smaller $\bf{I}_{\text{ser}}$ or $\bf{I}_{\text{ag}}$ implies that the risk of inspecting private information via word relationships by the server or other clients is decreased. The tables in Figure.~\ref{fig:wv-privacy} empirically verify that setting $d_1=2$ can lead to a stricter privacy protection.

	\subsubsection{Model Architectures}
	\label{sect:arch}
	Finally, we answer the question that why we only privatize a mlp classifier (Figure.~\ref{fig:method}) rather than the ``BiRNN-MLP'' or the whole model. We denote these three architectures (Arch.) as ``A'', ``B'' and ``C'' respectively. We remove the projection layers and the diversity term (i.e., $\lambda_2=0.0$), and only report the $\text{Ag}$ and $\text{Ap}$ under different settings of $\lambda_1$ in Table.~\ref{tab:ablation-arch}. Although ``B'' and ``C'' can obtain comparable $\text{Ag}$ results, the personalization results are worse than ``A''. Reasonably, the mlp classifier captures task-specific information and its number of parameters are not too large to overfit. Hence, only privatizing classifier will lead to better results. Additionally, this table also verifies the effectiveness of the added knowledge transfer term.

\section{Conclusion}
We carried on a preliminary research to MDSC with data privacy protection. We first sorted out the relationships between MTL and FL from the aspect of private-shared models, and correspondingly proposed a FL framework KTEPS for better model aggregation and personalization simultaneously under the Non-IID scenario. Additionally, for word embedding problems, we resorted to PDR and introduced KTEPS$^\star$ as a solution. Comprehensive experimental studies verified the superiorities of our methods. In summary, our work takes a small step towards FedMDSC. Considering fine-grained SC paradigms, cross-device FL settings, complex networks, task relationships, and advanced embedding compression methods are future works.

\newpage
\bibliographystyle{ACM-Reference-Format}
\bibliography{fedsent}

\end{document}